\documentclass[10pt,twocolumn,letterpaper]{article}

\usepackage[pagenumbers]{iccv} % To force page numbers, e.g. for an arXiv version

\newcommand{\TODO}[1]{\textbf{\color{red}[TODO: #1]}}
\renewcommand{\TODO}[1]{}

\definecolor{iccvblue}{rgb}{0.21,0.49,0.74}
\usepackage[pagebackref,breaklinks,colorlinks,allcolors=iccvblue]{hyperref}

\title{\vspace{-0.6cm} KernelFusion: Assumption-Free Blind Super-Resolution via Patch Diffusion}

\author{
    \parbox{\linewidth}{\centering
        \normalfont
        \vspace{-0.5em}
        \hspace{-2.75em} 
        Oliver Heinimann\textsuperscript{1*} \hspace{0.4em}
        Assaf Shocher\textsuperscript{2*} \hspace{0.4em} 
        Tal Zimbalist\textsuperscript{1} \hspace{0.4em}
        Michal Irani\textsuperscript{1} \hspace{0.4em}  \\
        \vspace{0.4em} 
        \small\textsuperscript{1}Weizmann Institute of Science \quad \quad \quad \quad \textsuperscript{2}NVIDIA Research 
    }
}

\usepackage{multirow}

\usepackage{amsmath}      % For mathematical notation
\usepackage{algorithm}    % For the algorithm environment
\usepackage{algpseudocode} % For the pseudocode formatting

\begin{document}
 % \maketitle

\twocolumn[{
\vspace{-0.5cm}
  \maketitle
% \vspace{0.1cm}
    % \hspace{-1.1cm}
    \centering%\vspace{-0.1cm}
    \includegraphics[width=0.9\linewidth]{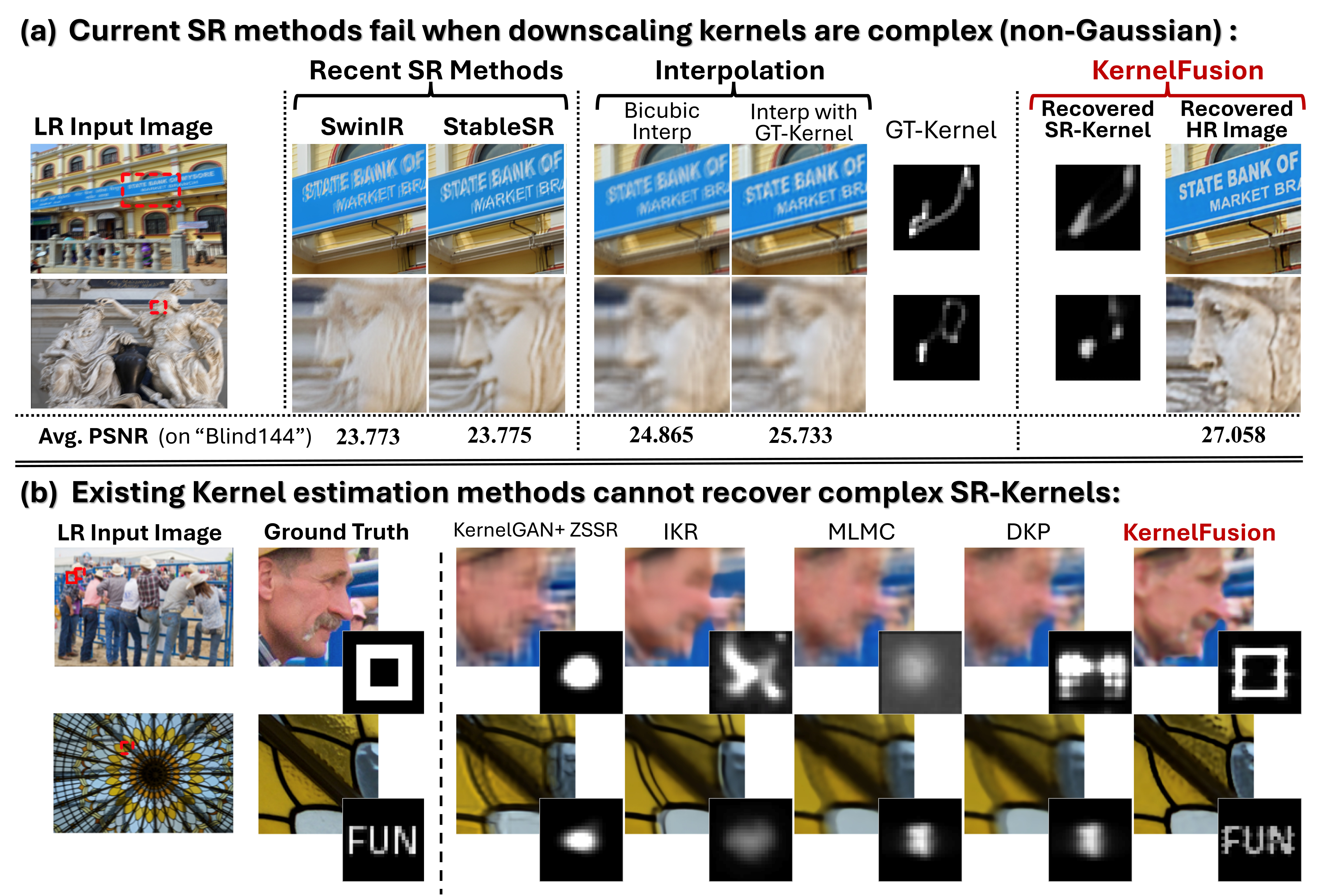}
  \vspace*{-0.3cm}
    \captionof{figure}{
    \small \textbf{The importance of an accurate SR-Kernel.}  {{{(A) {SotA SR-methods fail on complex downscaling kernels outside their training distribution}}},  performing even worse than interpolation on such kernels.}
    {{(B) {Existing SR-kernel estimation methods cannot handle complex downscaling kernels.}}} KernelFusion is the only method capable of estimating arbitrarily challenging SR-kernels.
    }
    \label{fig:teaser}
  \vspace{0.9em} 
}]

%%%% Abstract
\begin{abstract}
Traditional super-resolution (SR) methods assume an ``ideal'' downscaling SR-kernel (e.g., bicubic downscaling) between the high-resolution (HR) image and the low-resolution (LR) image. Such methods fail once the LR images are generated differently. 
Current blind-SR methods aim to remove this assumption, but are still fundamentally restricted to rather simplistic downscaling SR-kernels (e.g., anisotropic Gaussian kernels), and fail on more complex (out of distribution) downscaling degradations. 
However, using the correct SR-kernel is often more important than using a sophisticated SR algorithm.
In ``KernelFusion'' we introduce a zero-shot diffusion-based method 
that makes no assumptions about the kernel. Our method recovers the unique image-specific SR-kernel directly from the LR input image, while simultaneously recovering its corresponding HR image.
KernelFusion exploits the principle that the correct SR-kernel is the one that  maximizes patch similarity across different scales of the LR image. 
We first train an image-specific patch-based diffusion model on the single LR input image, capturing its unique internal patch statistics. 
We then reconstruct a larger HR image with the same learned patch distribution, while simultaneously recovering the correct downscaling  SR-kernel that maintains this cross-scale relation between the HR and LR images. Empirical results show that KernelFusion vastly outperforms all SR baselines on complex downscaling degradations, where existing SotA Blind-SR methods fail miserably. By breaking free from predefined kernel assumptions, KernelFusion pushes Blind-SR into a new assumption-free paradigm, handling downscaling kernels previously thought impossible.
\renewcommand{\thefootnote}{\fnsymbol{footnote}}
\footnotetext[1]{Equal contribution}
\newpage
\end{abstract}

%%%%%%%%%%%%%%%%%%%%%%%%%%%%%%%%%%%%%%%%%%%%%%%%%%%%%%%%%%%%%%%%%%%%%%%%%%%%%%
%%                            Introduction                                  %%
%%%%%%%%%%%%%%%%%%%%%%%%%%%%%%%%%%%%%%%%%%%%%%%%%%%%%%%%%%%%%%%%%%%%%%%%%%
\section{Introduction}
\label{sec:intro}
Super-resolution (SR) is an inverse problem of recovering a high-resolution (HR) image 
from its low-resolution (LR) counterpart, given by
\vspace{-0.225cm}
\begin{equation}
\label{eq: SR equation}
I_{LR} = \left( I_{HR} * k_s \right) \downarrow_s,
\end{equation}
where \(k_s\) is the downscaling kernel (also known as SR kernel) and
\(\downarrow_s\) denotes subsampling by a scale factor \(s\). 
Traditional SR methods have achieved impressive 
results ~\cite{dong2015srcnn, lim2017edsr, kim2016accurate, Zhang2018RCAN, zhang2018residual, saharia2022image, li2024blinddiff} under the assumption that $k_s$ is a global, known kernel (e.g.~bicubic with antialiasing), but this is rarely the case. 
The SR-kernel tends to be \emph{image-specific}; it is affected not only by sensor optics, but also by camera motion, subtle hand movements, and other factors. Evidently, these methods perform poorly in any scenario other than synthetic data specifically created using the assumed kernel. 
In fact, it was shown~\cite{levin2009, efrat2013accurate}, that the \emph{accuracy of the SR-kernel is often more critical for obtaining good SR}, than the image prior or the choice of SR algorithm used.

Blind-SR methods have emerged to address this limitation. Some approaches aim to explicitly estimate the unknown kernel
 (e.g.,~\cite{michaeli2013, kernelgan2019, mlmc}), whereas others 
represent the SR-kernel implicitly \cite{Gu2019IKC, realdan, Huang2020DAN, Kim2021KOALA, ikr}, or aim to design networks that are robust to kernel variations 
(e.g.,~\cite{liang2021swinir, wang2024sinsr, stablesr, Luo2022DCLS, luo2025kernel,lin2024diffbir}). 
However, existing Blind-SR methods are fundamentally limited: They can only super-resolve well LR images which were downscaled by simple, low-pass-filter kernels (e.g., (an)isotropic Gaussians), and fail on more complex downscaling kernels, which are 
outside their training distribution. 
In fact, \emph{for LR images obtained by non-Gaussian downscaling kernels, SOTA Blind-SR methods perform worse than simple interpolation.} (see Fig.~\ref{fig:teaser}a and Sec.~\ref{sec:kernel}). 

In ``KernelFusion'' we introduce a zero-shot diffusion-based method that makes no assumptions (explicitly or implicitly) about the downscaling kernel. 
Our method recovers the unique image-specific SR-kernel directly from the LR input image, while \emph{simultaneously} recovering its corresponding HR image.
KernelFusion exploits the principle (presented by~\cite{michaeli2013} and used in~\cite{kernelgan2019}), that the correct SR-kernel is the one that also maximizes patch similarity across different scales of the LR image. 
More specifically, we first train an \emph{image-specific} patch-based diffusion model on the single LR input image, capturing its unique internal patch statistics. 
We then freeze the patch-diffusion model, and use it to reconstruct a larger HR image with the same learned patch distribution. 
After each HR  diffusion denoising step, we iteratively recover the estimated HR image while simultaneously estimating the downscaling SR-kernel that maintains this cross-scale relation between the HR and LR images. 
Empirical results show that KernelFusion vastly outperforms all SR baselines on complex downscaling degradations, where existing SotA Blind-SR methods fail miserably. 

The success of KernelFusion in handling complex downscaling kernels (where all previous SR methods fail), stems from the following critical design choices: 
\vspace*{0,1cm} \\
\noindent
1. Being a zero-shot estimation method which trains \emph{internally} on the LR input image only, KernelFusion is not bound by any external training  distribution, hence can handle any type of downscaling kernel. There is no notion of ``out-of-distribution'' kernels, which \emph{externally-trained} Blind-SR methods suffer from  (see Fig.~\ref{fig:teaser}a).\vspace*{0,1cm}  \\
\noindent
2. Previous zero-shot SR-kernel methods~\cite{michaeli2013, kernelgan2019} estimated the kernel only. They require a separate independent SR algorithm to super-resolve the LR image with their recovered kernel (e.g., using ZSSR~\cite{Shocher2018ZSSR} or SRMD \cite{Zhang2018SRMD}, which can receive a user-specified kernel). Such a 2-step process suffers from accumulated errors and inconsistencies between the estimated kernel and the estimated HR image.  In contrast, KernelFusion \emph{simultaneously} estimates both the SR-kernel and the corresponding HR image \emph{in a consistent manner}.
\vspace*{0,1cm} \\
\noindent
3. The explicit kernel estimation methods of~\cite{kernelgan2019,mlmc,ikr, yang2024dynamic_kernel_prior} seem to recover well only specific types of kernels (Gaussians and motion lines). We suspect that this limitation stems from the implicit-bias of the CNN and MLP architectures, which tend to produce smooth outputs (as also observed in~\cite{tancik2020fourier}). In contrast, the kernel-estimation component in KernelFusion employs an \emph{Implicit Neural Representation} (INR) architecture, which allows for the recovery of complex \emph{non-smooth} downscaling SR-kernels (see Fig.~\ref{fig:teaser}).

By breaking free from predefined kernel assumptions, KernelFusion pushes Blind-SR into a new, assumption-free paradigm — handling out-of-distribution  (non-Gaussian) downscaling kernels, which were previously out of reach.

%%%%%%
\vspace*{0.1cm}
\noindent\textbf{Our contributions are therefore as follows:}
\begin{itemize}[noitemsep]
\item \emph{KernelFusion} is the first deep Blind-SR method capable of recovering arbitrary downscaling SR-kernels.
\item \emph{KernelFusion} obtains state-of-the-art SR results on challenging downscaling degradations, where leading SR methods fail.
\end{itemize}

\begin{figure*}
    \centering
    \includegraphics[width=1\linewidth]{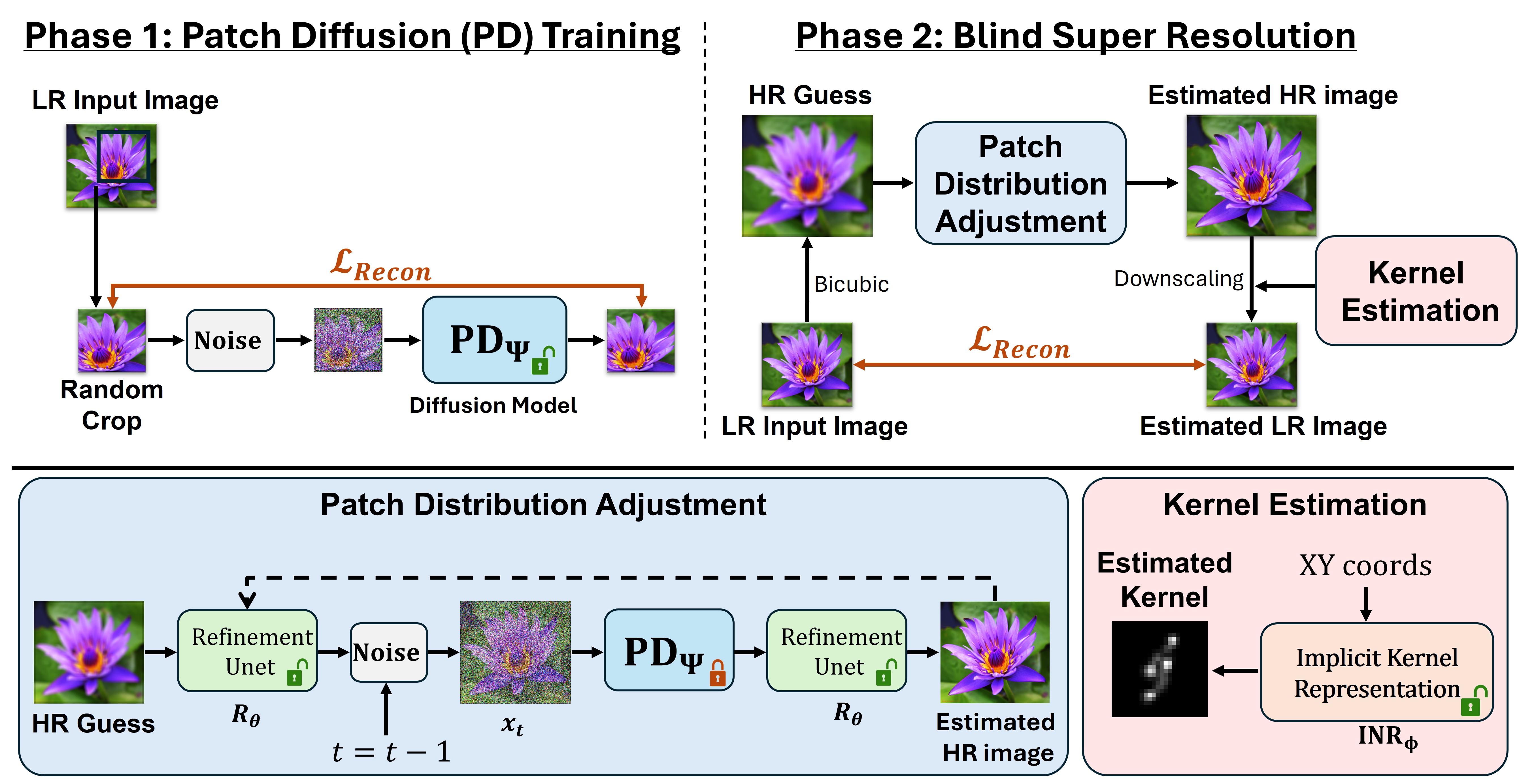}
    \caption{\textbf{Method Overview.} { \ Our approach consists of 2 stages: \  \underline{\bf Phase 1}: We train a diffusion model (PD) to learn the patch distribution of a single image. \ \underline{\bf Phase 2}: We perform blind SR and kernel estimation simultaneously. In particular, we use the trained PD to shift the HR guess toward the patch distribution of the LR input. A refinement U-Net and an implicit kernel representation model are trained jointly under a consistency loss, ensuring that convolving the estimated HR image with the learned kernel reproduces the original LR image.}
}
    \label{fig:method overview}
\end{figure*}

%%%%%%%%%%%%%%%%%%%%%%%%%%%%%%%%%%%%%%%%%%%%%%%%%%%%%%%%%%%%%%%%%%%%%%%%%%%%%%
%%                            Related Work                                %%
%%%%%%%%%%%%%%%%%%%%%%%%%%%%%%%%%%%%%%%%%%%%%%%%%%%%%%%%%%%%%%%%%%%%%%%%%%%%%%

\section{Related Work}
\label{sec:related_work}
We start by describing three main types of Blind-SR \cite{liu2022blind}. Finally, we review diffusion models and their use for inverse problems, which is related to our method.

\noindent
\textbf{Blind-SR by training on synthetic degradations:} One type of Blind-SR involves training on data with synthetic degradations  out of a predefined distribution of degradations. They can be applied to data close to that distribution and achieve visually pleasing results. Among them are SwinIR \cite{liang2021swinir}, Real-ESRGAN \cite{Wang2021RealESRGAN} and many more \cite{conde2022swin2sr, Zhang2021BSRGAN, Jo2021AdaTarget, Luo2022DCLS, lin2024diffbir, sun2024rfsr, wu2024seesr, wei2020component, yang2024pixel, zhang2024real}. These methods are inherently restricted by their training distribution, thus fail on LR images generated by out-of-distribution downscaling kernels (e.g., non-Gaussian kernels).

\noindent
\textbf{Blind-SR with latent kernel representation:} other methods represent degradations by a latent features vector. These methods perform SR and refinement of the degradation features, often based on alternating SR and latent kernel estimation as was first shown by IKC \cite{Gu2019IKC}, later improved by unfolding this alternation with DAN \cite{Huang2020DAN, realdan} and many more \cite{Luo2022DCLS, Kim2021KOALA, Zhang2020USRNet, sohl2015deep}. The kernel representation is data-driven, hence also based on the synthetic degradations used at training. This makes them fail, just like the robustness based methods, on images downscaled by kernels out of their training distribution.

\noindent
\textbf{Blind SR-Kernel Estimation:}
Acknowledging the importance of an accurate SR-kernel, some approaches  aim to explicitly estimate the unknown SR-kernel
directly from the LR image (e.g.,~\cite{michaeli2013, kernelgan2019, mlmc, ikr, yang2024dynamic_kernel_prior}).
Notably, \cite{michaeli2013} was the first to observe that the optimal SR-kernel is the one that maximizes the similarity of small patches \emph{across} different scales of the LR image, and accordingly used cross-scale patch nearest-neighbors to estimate the SR-kernel.
KernelGAN~\cite{kernelgan2019} further used this principle within deep learning, showing that the SR-kernel can be estimated by training an image-specific GAN on the LR image. However, the zero-shot methods of~\cite{michaeli2013, kernelgan2019}  are pure kernel estimation methods, and require a separate followup algorithm to perform the SR step on the LR image (e.g., using ZSSR~\cite{Shocher2018ZSSR} or SRMD \cite{Zhang2018SRMD}), which can  receive also a user-specified kernel as an input. This restricts their applicability.
Moreover, the explicit kernel estimation methods of~\cite{kernelgan2019,mlmc,ikr, yang2024dynamic_kernel_prior} can only recover well specific types of kernels (Gaussians and motion lines 
-- see examples in Fig.~\ref{fig:teaser}).

\noindent
\textbf{Diffusion Models and Inverse Problems:}  
Diffusion probabilistic models \cite{sohl2015deep, ho2020denoising} have become a powerful tool for modeling complex image distributions. More recently, under the Deep Internal Learning regime \cite{Shocher2018ZSSR, gandelsman2019double, ulyanov2018deep, shocher2018ingan, shaham2019singan, granot2022drop} diffusion-based approaches have been adapted for the single-image setting \cite{Nikankin2023SinFusion, Wang2022SinDiffusion, Kleiner2023SinDDM}. Additionally, diffusion models have shown promise in solving inverse problems such as deblurring and super-resolution \cite{kawar2022denoising, Chung2022DPS}.
Recent works further enhance these approaches by incorporating data-consistency constraints via null-space projections \cite{Wang2023DDNM} or by adding back-projection steps \cite{Hui2024ShortcutDiff}.  
These techniques enable zero-shot restoration without retraining, and they demonstrate that iterative diffusion-based refinement can recover high-quality images under unknown degradations. Additional work on patch-based diffusion models \cite{Hu2024PatchDiffusion,Hu2024PaDIP} further exploits local image statistics for improved detail recovery.
Due to their stochastic nature, diffusion models often struggle to strictly adhere to measurement constraints. Moreover, pretrained diffusion models may exhibit a distribution mismatch between the training data and the observed measurements, necessitating careful adaptation \cite{Hu2024PaDIP}.

%%%%%%%
\section{The Importance of an Accurate SR-Kernel}
\label{sec:kernel}

The accuracy of the SR-kernel is critical for achieving high-fidelity HR image reconstruction, often playing a more crucial role than the image prior or SR method itself~\cite{efrat2013accurate, levin2009}.
The SR process fundamentally relies on inverting the degradation introduced by downsampling, which is dictated by the underlying blur kernel. 
If the kernel is inaccurate or out of distribution, even advanced SR models risk amplifying artifacts and producing unrealistic results.

This property is shown in Fig.~\ref{fig:teaser}: Two state-of-the-art algorithms (SwinIR and StableSR), both trained with blind SR degradations, fail to accurately upscale LR images which weredownscaled by non-Gaussian kernels. In contrast,  simple interpolation leads to remarkably better visual results. 
We further confirm this observation quantitatively on 2 datasets with hundreds of LR images downscaled by a variety of \emph{non}-gaussian kernels (``Blind144'' \& ``DIV2KFK'' -- see Sec.~\ref{sub:datasets}). These results are summarized in Table~\ref{tab:backprojection}.
The table shows that SotA Blind-SR methods perform \emph{worse} than simple bicubic interpolation on non-gaussian kernels (which are outside their training distribution). Moreover, applying more sophisticated interpolation\footnote{Interpolation with a kernel is obtained by backprojection~\cite{irani1991improving} 
with the estimated Pseudo-Inverse~\cite{Moore1920pinv, Penrose1955pinv} of the kernel.} 
with the ground-truth (GT) kernel provides an additional large improvement by $+$1bB over SotA Blind-SR methods on such kernels.

\begin{table}[ht]
\centering
\resizebox{0.48\textwidth}{!}{%
\hspace*{-0.25cm}
\begin{tabular}{lcccc}
\toprule
\multirow{2}{*}{\textbf{\Large Methods}} & \multicolumn{2}{c}{\textbf{\Large Blind144}} & \multicolumn{2}{c}{\Large \textbf{DIV2KFK}} \\
\cmidrule(lr){2-3} \cmidrule(lr){4-5}
 & \textbf{\large PSNR$\uparrow$} & \textbf{\large SSIM$\uparrow$} & \textbf{\large PSNR$\uparrow$} & \textbf{\large SSIM$\uparrow$}\\
\midrule
\Large Bicubic interpolation & \large  24.865 & \large 0.637 & \large 24.101 & \large 0.639 \\
\midrule
\Large Interpolation with the GT kernel & \large  25.733 & \large 0.659 & \large 25.057 & \large 0.666 \\
\midrule
\Large SotA: DCLS-SR \cite{Luo2022DCLS}/ DPSR \cite{zhang2019deep}  & \large 24.808 & \large 0.633 & \large 23.997 & \large 0.637 \\
\bottomrule
\end{tabular}%
}
%\vspace*{-0.3cm}
\caption{
SotA Blind-SR methods perform \emph{worse than interpolation} on LR images downscaled by non-gaussian kernels
%Mean PSNR and SSIM computed on 
(Blind144 \& DIV2KFK are two such Blind-SR datasets - see Sec.~\ref{sub:datasets}
\vspace{-0.25cm}).
}
\label{tab:backprojection}
\end{table}

%%%%%%%%%%%%%%%%%%%%%%%%%%%%%%%%%%%%%%%%%%%%%%%%%%%%%%%%%%%%%%%%%%%%%%%%%%%%%%
%%                            Method                                        %%
%%%%%%%%%%%%%%%%%%%%%%%%%%%%%%%%%%%%%%%%%%%%%%%%%%%%%%%%%%%%%%%%%%%%%%%%%%%%%%
\section{Method: ``KernelFusion'' for Blind-SR} 
Our method builds on the principle that the correct SR kernel is the one that best preserves the image's patch distribution across scales. In a first step, a patch-based diffusion model is trained on the LR input image, learning its patch distribution. In a second step, we perform super resolution by iteratively improving our HR guess and estimating the SR-kernel during the reverse diffusion process. Fig.~\ref{fig:method overview} provides an overview over our approach.

\subsection{Phase 1: Training Patch-Diffusion (PD)}
Our patch-diffusion (PD) model aims to learn the patch distribution of the  LR input image $I_{LR}$. As diffusion models are excellent distribution learners, we base our model on the standard denoising probabilistic diffusion model (DDPM)~\cite{ho2020denoising}, predicting velocity $v_t$ as proposed in \cite{salimans2022progressive}.
We train it to denoise solely the single LR input image noised with various noise magnitudes, according to a standard DDPM schedule. More specifically, the input is a diffused LR input image to random time step $t\in [1, T]$ while the target is the velocity (from which one can easily derive the clean original input image using a closed form):
\vspace{-0.3cm}
\begin{align}
\Psi &= argmin_\psi \Big\| PD_\psi\left(x_t\right) - v_t \Big\|_2^2,
\end{align}
where $x_t$$=$$\sqrt{\bar{\alpha}_t} I_{LR}$$+$$\sqrt{1-\bar{\alpha}_t}$, \  
$v_t$$=$$\sqrt{\bar{\alpha}_t} \epsilon$$-$$\sqrt{1-\bar{\alpha}_t}I_{LR}$, $\Psi$ are the parameters of our PD model, $\epsilon$ is a standard normal white noise, $\bar{\alpha_t}$ is the standard DDPM coefficient. 

Learning the patch distribution of a single image requires some adjustments: 
We took inspiration from \cite{Nikankin2023SinFusion} using pure CNN without any global layers (such as attention). However, since our method is based on much smaller patches, the receptive field needs to be restricted even further. Thus we exchange the backbone with a simple convolutional network with no strides, introducing a receptive field of 15$\times$15 pixels.
%px.
This allows training PD on random 64$\times$64 image crops.
Further design choices and implementation details can be found in Sec.~\ref{subsec:Implementation Details}.

\subsection{Phase 2: Reverse Diffusion at High Resolution}
Once PD is trained, it has learned the patch distribution of the LR input image. Next, we perform \emph{simultaneous} SR and kernel estimation, based on the reversed sampling process of PD. This constraints the produced HR image to have the same patch distribution as $I_{LR}$ (which it should have~\cite{michaeli2013}). This combined approach has a major benefit: A better HR prediction $\hat{x}_0$ leads to an improved kernel prediction, and vice versa. Algorithm~\ref{alg:ddpm_reverse} provides pseudo-code of the approach, which is further detailed below.

Our method takes as an initial input a bicubically upscaled version of the LR image $I_{LR}$. This bicubic guess is noised by $T_{nd}$ steps using the diffusion noise schedule. 
Subsequently, we would like to optimize $x_0$ at each timestep $t$, such that the downscaled version  $\hat{x}_0\downarrow_s$ is consistent with the input image $I_{LR}$. Instead of directly optimizing $x_0$, we chose to implicitly optimize $x_0$ via a U-Net (see Fig.~\ref{fig:method overview}). The DIP U-Net~\cite{ulyanov2018deep} imposes a \emph{global} image prior on the output.
%This allows us to generalize better while making use of the  architecture prior as known in \cite{ulyanov2018deep}.
The PD step used to predict $\hat{x}_0$ alone does not inherently preserve global structure, especially at high $t$, due to the 15$\times$15 local receptive field of PD.
We solve this problem by applying a DIP U-Net twice: First, we apply the U-Net to $x_0$ from the previous timestep $t+1$ and reconstruct the required $x_t$ from it. Second, we apply the U-Net after denoising $x_t$ using our patch-diffusion model $PD_\psi$ to the predicted $x_0$ at timestep $t$.
Optimizing the U-Net using our LR consistency loss is hence key to maintain global structure. 
This setup enables joint training of both the DIP UNET and the kernel estimation network (see Sec.~\ref{subsec:kernel estimation}) with a single loss.
Since both networks are trained from scratch, we apply $n_{iter}$ gradient steps at each timestep $t$. The networks are iteratively refined at every $t$, leveraging their gradual improvements to enhance the prediction of $\hat{x}_0$.

\begin{algorithm}[t]
    \caption{Reverse Diffusion Process in KernelFusion}
    \label{alg:ddpm_reverse}
    \begin{algorithmic}[1]
        \Require  
        \begin{itemize}  
            \item[] 
            \item Pretrained velocity prediction model: $\mathrm{PD}_\psi$  
            \item Number of noise/denoise timesteps: $T_{nd}$  
            \item Noise schedule: $\{\beta_t\}_{t=1}^{T}$ with $\alpha_t = 1 - \beta_t$
            \item Initial noised bicubic guess: 
            \item[] $x_{T_{nd}} = \sqrt{\bar{\alpha}_{T_{nd}}}(I_{LR} \uparrow_{bic,s}) + \sqrt{1-\bar{\alpha}_{T_{nd}}}\epsilon$ 
            \item[] with $\epsilon \sim \mathcal{N}(0, I)$
            \item U-Net network: $\mathrm{R}_\theta$  
            \item SIREN network: $\mathrm{INR}_\phi$  
            \item Kernel coordinate grid: $g$  
            \item Number of optimization steps per $t$: $n_{iter}$  
            \item Learning rate: $\gamma$  
        \end{itemize} 
        \Ensure Generated sample $x_0$.
        \For{$t = T_{nd}$ to $1$}
            \State $\xi \sim \mathcal{N}(0, I)$ \Comment{Sample noise}
            \For{$i = 1$ to $n_{iter}$}
                \State $\hat{k}_\phi = \mathrm{INR}_\phi(g)$ \Comment{get kernel}
                \State $\hat{x}_{0,t+1,\theta} = \mathrm{R}_\theta(x_{0,t+1})$ \Comment{get optimized $x_0$}
                \State $\hat{x}_t = \mu_{t+1}(\hat{x}_{0,t+1,\theta}, x_{t+1}) + \sigma_{t+1}\xi$
                \State $\hat{v}_\theta = \mathrm{PD}_\psi(\hat{x}_{t,\theta}, t)$ \Comment{denoising step}
                \State $\hat{x}_{0,t} = \sqrt{\bar{\alpha_t}} \cdot \hat{x}_{t,\theta}  - \sqrt{1 - \bar{\alpha_t}} \cdot \hat{v}_\theta$
                \State $\hat{x}_{0,t,\theta} = \mathrm{R}_\theta(\hat{x}_{0,t})$ \Comment{get optimized $x_0$}
                \State $\hat{x}_{0,t}\downarrow_s = (\hat{x}_{0,t} * \hat{k}_\phi)\downarrow_s $ \Comment{Downscale step}
                \State $\hat{x}_{0,t+1}\downarrow_s = (\hat{x}_{0,t+1} * \hat{k}_\phi)\downarrow_s $ \Comment{Downscale step}
                \State $\mathcal{L} =
                    \begin{aligned}[t]
                        &\left\| I_{LR} - \hat{x}_{0,t}\downarrow_s \right\|_2^2 + \left\| I_{LR} - \hat{x}_{0,t+1}\downarrow_s \right\|_2^2 \\
                        &+ \mathrm{COM}(\hat{k}_\phi)
                    \end{aligned}$
                \State $\theta = \theta - \gamma \nabla_{\theta}(\mathcal{L} )$ \Comment{Step for U-Net}
                \State $\phi = \phi - \gamma \nabla_{\phi}(\mathcal{L} )$ \Comment{Step for INR}
            \EndFor
            
            \State $\mu_t = \beta_t \frac{\bar{\alpha}_{t-1}}{1 - \bar{\alpha}_t}\hat{x}_{0,\theta, \psi} + (1-\bar{\alpha}_{t-1})\frac{\alpha_t}{1-\bar{\alpha}_t}\hat{x}_{t,\theta}$
             \State $\zeta \sim \mathcal{N}(0, I)$ \Comment{Sample noise}
             \State $x_{t-1} = \mu_t + \sigma_t \zeta$ \Comment{Add stochasticity, $\sigma_0 = 0$}
        \EndFor
        \State \Return $x_0$
    \end{algorithmic}
\end{algorithm}

\subsection{Kernel Estimation using INR}
\label{subsec:kernel estimation}
Estimating a SR-kernel means solving for $k_s$ in Eq.~(\ref{eq: SR equation}).
Instead of directly solving for $k_s$, we chose to represent the kernel via an Implicit Neural Representation (INR) network, which allows us to represent the kernel \emph{continuously}~\cite{shocher2020discrete}, while controlling its level of regularization. Specifically, we took inspiration by the SIREN architecture~\cite{sitzmann2020implicit} which is known for its ability to also represent high frequency functions. Specifically, the sinusoidal activations enable the network to capture fine-grained structures without introducing over-smoothing, which is critical for accurately estimating complex downscaling kernels $k_s$.

\subsection{Implementation Details}
\label{subsec:Implementation Details}
\paragraph{PD Architecture} To ensure a small receptive field, we chose to exchange the global-receptive-field U-net of the original DDPM architecture with small, fully convolutional neural network. We use one block consisting of two 3$\times$3 convolutions, followed by six blocks 3$\times$3 + 1$\times$1 convolutions. This results in a theoretical receptive field of 15$\times$15 pixels. 
The diffusion model is trained with $T$$=$$1000$  time steps.
Further implementation details are found in the supplementary material (Sec.~\ref{supp:sec:technical details}).

%%%%%%%%%%%%%%%%%%%%%%%%%%%%%%%%%%%%%%%%%%%%%%%%%%%%%%%%%%%%%%%%%%%%%%%%%%%%%%
%%                            Results                                       %%
%%%%%%%%%%%%%%%%%%%%%%%%%%%%%%%%%%%%%%%%%%%%%%%%%%%%%%%%%%%%%%%%%%%%%%%%%%%%%%
\section{Results}
\label{sec:results}

\begin{figure*}[ht]
    \centering
    \includegraphics[width=\linewidth]{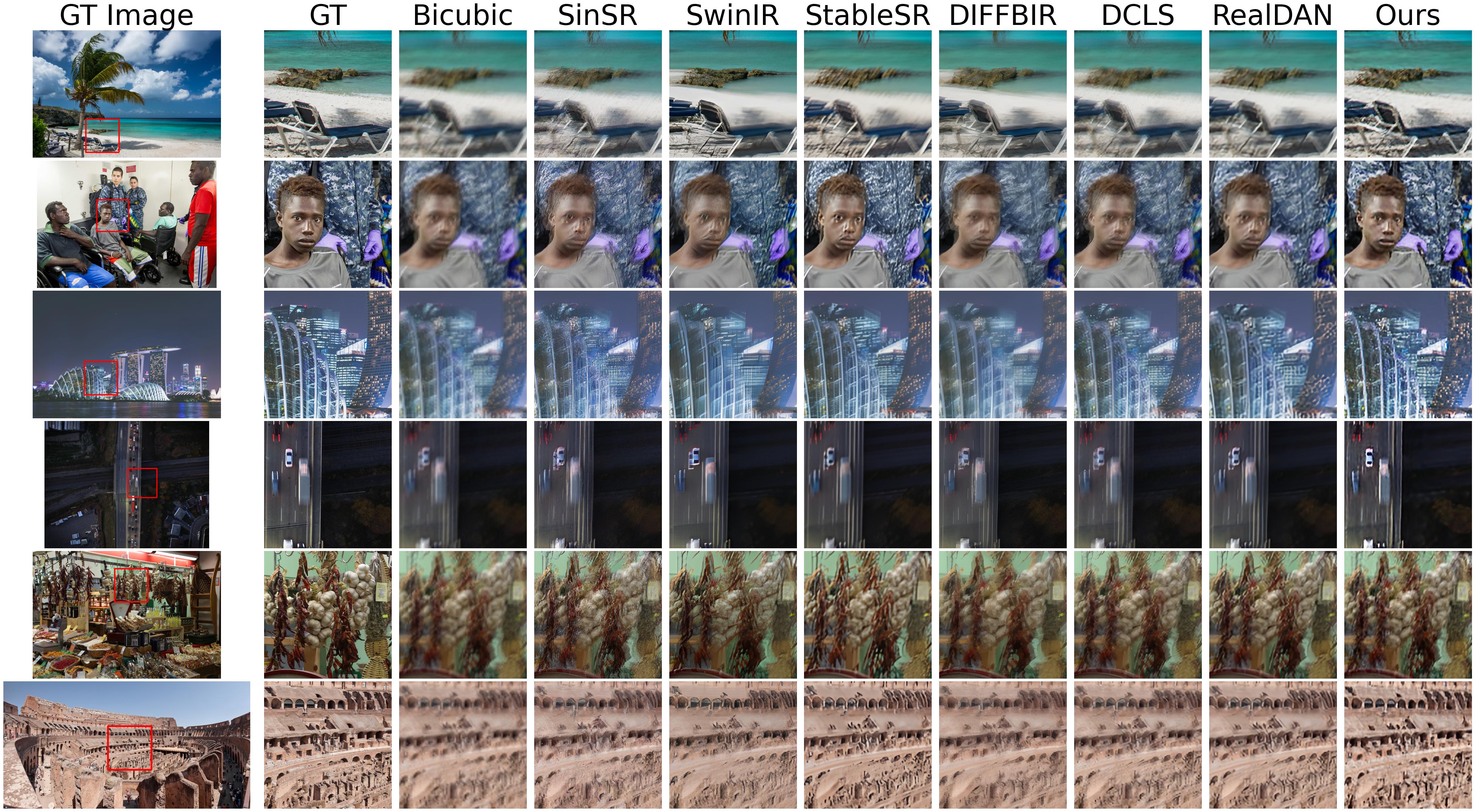}% Placeholder image
    \vspace*{-0.2cm}
        \caption{\textbf{Blind-SR comparison} on the DIV2KFK dataset (4× upscaling). Each row corresponds to a different degraded image from DIV2KFK, while each column shows the output of a different method at a 4× upscaling factor. Notably, our method reduces doubling artifacts in structured patterns (e.g., aerial road scene, 4th row), demonstrating its effectiveness in restoring fine details and mitigating motion  effects. 
        }
    \label{fig:sr}
\end{figure*}

\begin{figure*}[ht]
    \centering
    \subfloat[]{\includegraphics[width=0.5\textwidth]{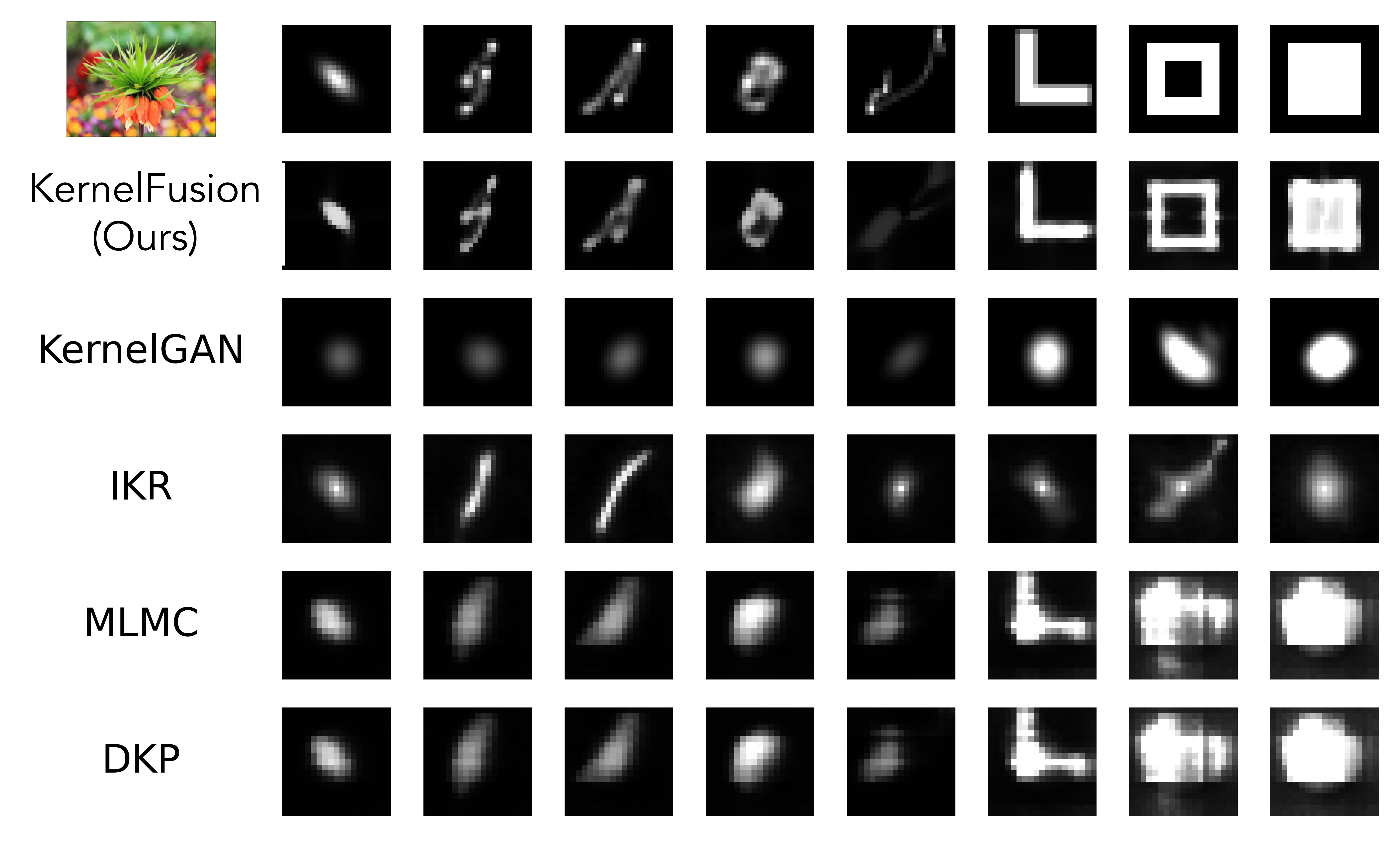}}
    \hfill % Add horizontal space between subfigures
    \subfloat[]{\includegraphics[width=0.5\textwidth]{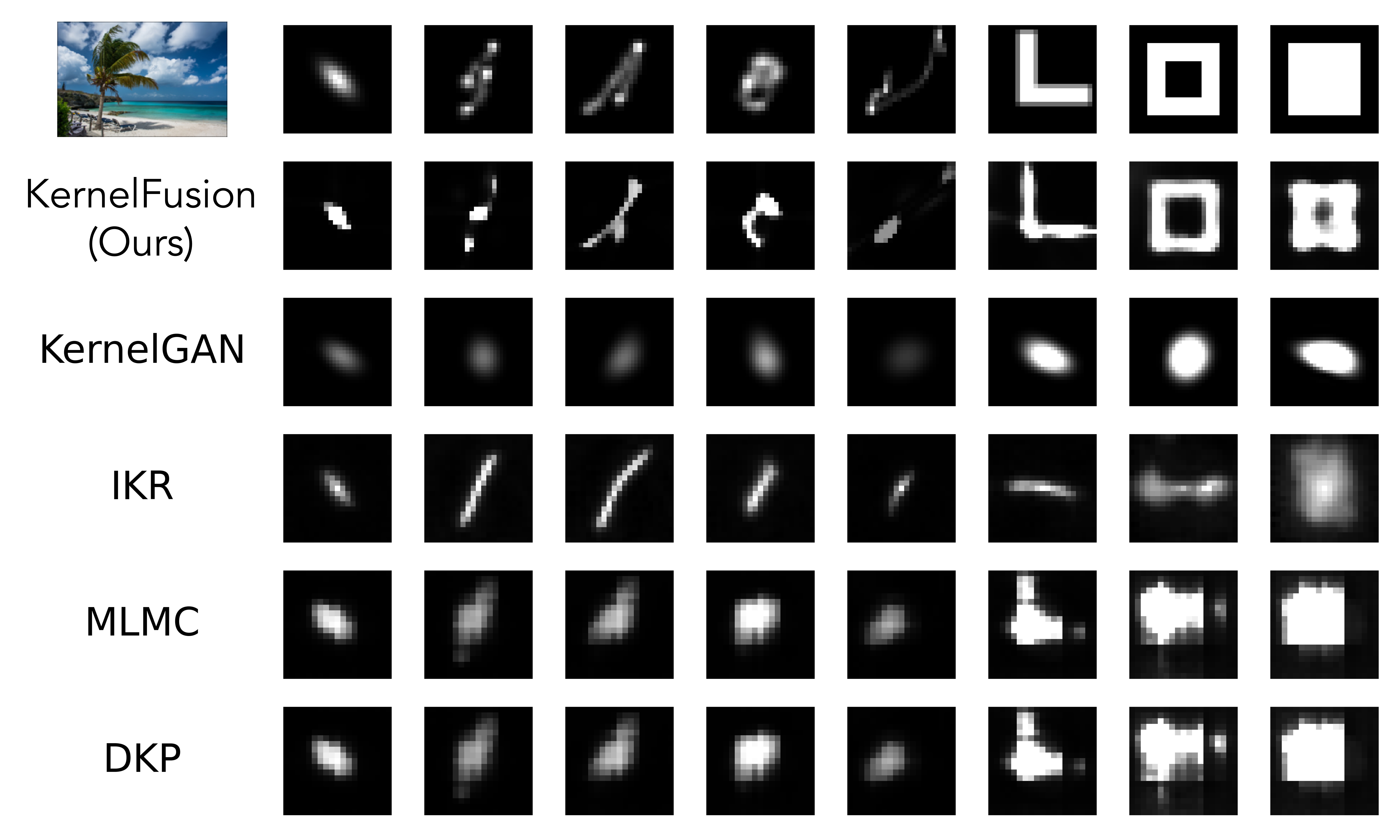}}
\vspace*{-0.4cm}
    \caption{\textbf{Comparison of estimated kernels from different Blind-SR methods.} The top row represents the ground-truth (GT) degradation kernels, while each subsequent row corresponds to the estimated kernels from different SR methods, including our approach, KernelGAN, IKR, MLMC, and DKP. Our method demonstrates superior flexibility in recovering complex, non-Gaussian degradations, accurately capturing kernel structures across a diverse range of degradations.  
    }
    \label{fig:est kernels vs competitors}
\end{figure*}

\begin{figure*}[ht]
    \vspace{-0.4cm}
    \centering
    \includegraphics[width=0.9\linewidth]{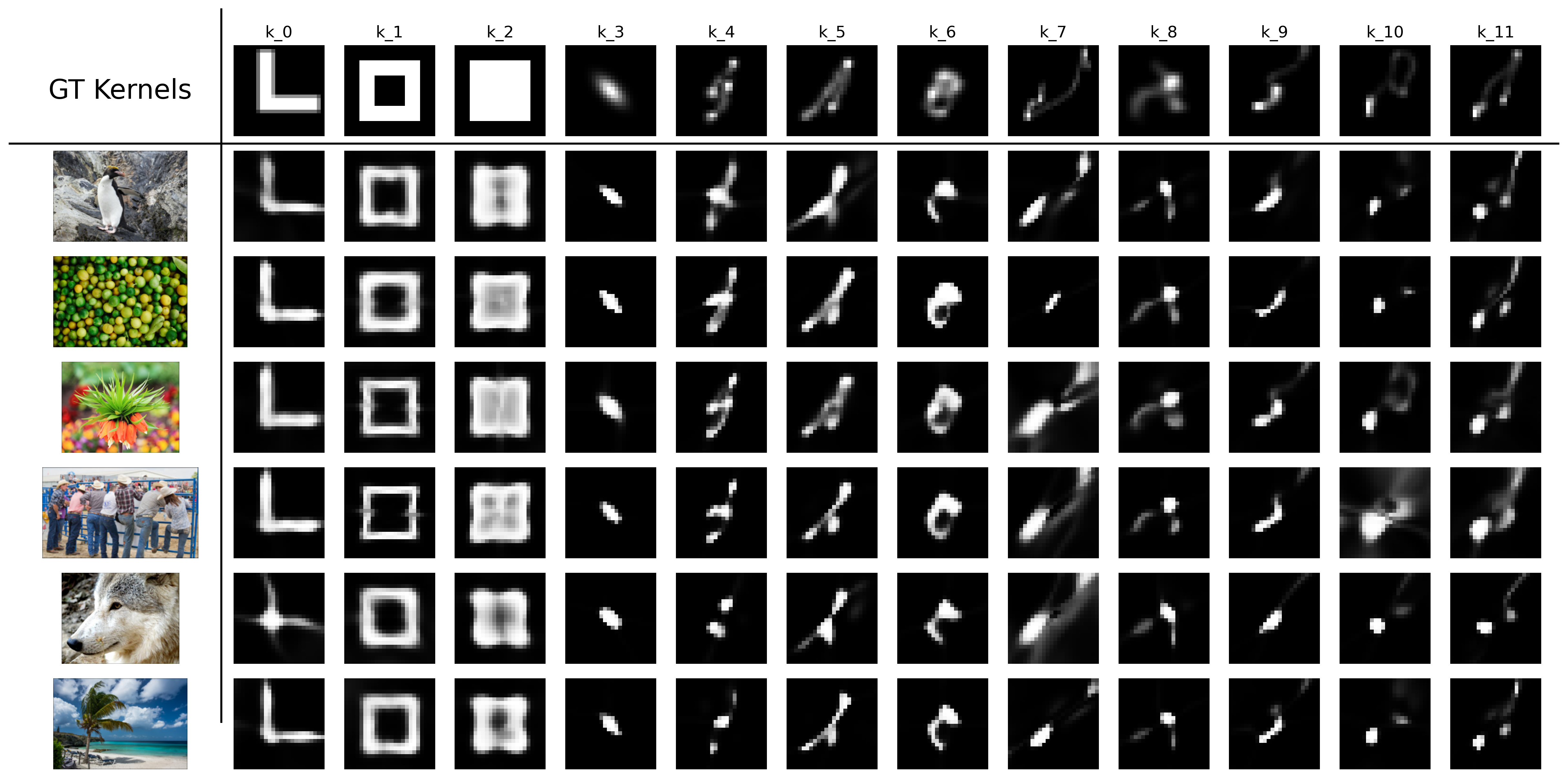}
    \vspace*{-0.3cm}
    \caption{\textbf{Kernel estimation results on Blind144}. The top row displays the 12 ground-truth (GT) degradation kernels, including real-world motion blur kernels from \cite{levin2009}, an anisotropic Gaussian kernel, and three synthetic non-natural kernels (L-shape, empty square, and filled square). The subsequent rows show our method’s estimated kernels for each of the 12 kernels applied to the first 12 images of the DIV2K validation set. Our approach successfully captures a diverse range of degradations, including complex structured kernels, demonstrating its robustness and adaptability in blind SR kernel estimation. 
    }
    \label{fig:kernel panel}
\end{figure*}

\begin{table*}[ht]
% \centering
\renewcommand{\arraystretch}{1.1} % Increase row height slightly for readability
\setlength{\tabcolsep}{4pt}       % Reduce column spacing
\fontsize{9pt}{11pt}\selectfont   % Reduce font size slightly
\hspace{-1cm}
\resizebox{1.1\textwidth}{!}{
\begin{tabular}{ll*{15}{c}} 
\toprule
\textbf{\large Dataset} & \textbf{\large Metric} & 
\shortstack{Bicubic\\interpolation} & 
\shortstack{SinSR\\\cite{wang2024sinsr}} & 
\shortstack{StableSR\\\cite{stablesr}} & 
\shortstack{SwinIR\\\cite{liang2021swinir}} & 
\shortstack{DPSR\\\cite{zhang2019deep}} & 
\shortstack{IKR\\\cite{ikr}} & 
\shortstack{RealDAN\\\cite{realdan}} & 
\shortstack{RealDAN \\GAN} & 
\shortstack{RealDAN \\specialized} & 
\shortstack{DCLS-SR\\\cite{Luo2022DCLS}\footnote{A follow-up work with an improved DCLS model has been published in \cite{luo2025kernel}, but the code has not been released at the time of submission.}} & 
\shortstack{DiffBIR\\\cite{lin2024diffbir}} & 
\shortstack{DKP\\\cite{yang2024dynamic_kernel_prior}} & 
\shortstack{MLMC\\\cite{mlmc}} & 
\shortstack{KernelGAN\\+ZSSR \cite{kernelgan2019}} & 
\shortstack{\normalsize \textbf{KernelFusion}\\ \textbf{\normalsize (ours)}}
\\
\midrule
\multirow{2}{*}{\large Blind144} 
  & PSNR$\uparrow$ & \underline{24.865} & 23.587 & 23.775 & 23.773 & 24.824 & 24.113 & 24.624 & 23.998 & - & 24.808 & 24.259 & 23.431 & 23.430 & 24.529 & \textbf{27.058} \\
  & SSIM$\uparrow$ & 0.637  & 0.582  & 0.625  & 0.616  & 0.637  & 0.630  & \underline{0.638}  & 0.619  & - & 0.633  & 0.599  & 0.612  & 0.612  & 0.633  & \textbf{0.715} \\
\midrule
\multirow{2}{*}{\large DIV2KRK} 
  & PSNR$\uparrow$ & 25.075 & 25.360 & 25.262 & 25.139 & 25.317 & 23.906 & 26.870 & 26.057 & \textbf{27.821} & \underline{27.150} & 25.431 & 23.127 & 23.122 & 25.895 & 26.747 \\
  & SSIM$\uparrow$ & 0.671  & 0.680  & 0.709  & 0.699  & 0.682  & 0.667  & 0.745  & 0.730  & \textbf{0.775} & \underline{0.748} & 0.668  & 0.629  & 0.629  & 0.703  & 0.714 \\
\midrule
\multirow{2}{*}{\large DIV2KFK} 
  & PSNR$\uparrow$ & \underline{24.101} & 22.887 & 23.077 & 23.070 & 23.977 & 23.352 & 23.941 & 23.439 & - & 23.886 & 23.546 & 22.531 & 22.535 & 23.617 & \textbf{26.155} \\
  & SSIM$\uparrow$ & 0.639  & 0.585  & 0.630  & 0.620  & 0.637  & 0.627  & \underline{0.644} & 0.628  & - & 0.634  & 0.606  & 0.603  & 0.604  & 0.629  & \textbf{0.711} \\
\bottomrule
\end{tabular}
}
\caption{\textbf{Quantitative comparison of methods on 4x SR across different Blind-SR datasets.} The 2 best results per dataset (row) are highlighted in \textbf{bold} and \underline{underline}. 
Note that on the 2 non-gaussian datasets (DIV2KFK and Blind144), KernelFusion is best (by a large margin), and Bicubic interpolation is second. \ \ 
(\underline{Comments:} (i)~``RealDAN specialized'' is a version of ``RealDAN'' specialized  for Gaussian kernels only, hence applied only to DIV2KRK. (ii)~The reported PSNRs are after excluding an image boundary of 5\%, due to boundary effects.)
}
\label{tab:combined_results_transposed}
\end{table*}

\subsection{Datasets}
\label{sub:datasets}
We evaluate our method on a variety of blind SR datasets:

\noindent 
\textbf{DIV2KRK \cite{kernelgan2019}:}
From DIV2K~\cite{div2k} validation set (100 images). LR samples are obtained by downscaling each image with a different random gaussian kernel, hence containing anisotropic Gaussians of varying sizes and orientations.

\noindent 
\textbf{DIV2KFK:}
Inspired by DIV2KRK, we created a new dataset based on the 8 real-life downscaling kernels measured by Levin~\cite{levin2009}, which were induced by small camera jitter during the shutter exposure time.
We call this dataset DIV2KFK (DIV2K-Fancy-Kernels).
In order to have reaonable blur levels for 4x SR, we resize the original kernels to size $24$$\times$$24$.
Each image of the DIV2K validation set is convolved with a randomly selected kernel and then subsampled by a scale factor of 4.

\noindent 
\textbf{Blind144:} 
Two factors affect the Blind-SR challenge: the image and the downscaling kernel. To further analyze and better understand SR results, we created the controlled Blind144 dataset. It is organized as a 12$\times$12 matrix comprising 12 images and 12 kernels. This setup allows us to examine the results of a single kernel across various images, as well as the outcomes of a single image across different kernels, thereby clarifying whether specific effects and behaviors stem from the image or the kernel. The images used are the first 12 images from DIV2K. For the selection of kernels we used the 8 empirical kernels as in DIV2KFK, and in addition 3 extreme stress-testing kernels and one anisotropic Gaussian. The purpose of the Gaussian kernel is to compare quality of results on the same image for in-distribution vs.\ out-of-distribution kernels.
The 3 extreme kernels contain an "L" shaped kernel, a full square with sharp edges, and an empty square. These allow stress-testing of kernel extraction. The full set of kernels is displayed in the top row of Fig.~\ref{fig:kernel panel}.

\subsection{Empirical Evaluation of SR}
%\subsection{Evaluation Method}
We evaluate the quality of the recovered HR images using peak signal-to-noise ratio (PSNR) and structural similarity index measurement (SSIM)~\cite{wang2004ssim}, with respect to the GT HR images.
We orient our evalution on the procedure of~\cite{kim2016accurate}:
The evaluation is executed on the luminance channel (YCBCR space).
An image boundary of $\sim$5\% is excluded in the PSNR/SSIM estimation, due to boundary effects.

Table~\ref{tab:combined_results_transposed} 
compares KernelFusion with a variety of SotA competitors.
Our method surpasses all SotA blind-SR methods by a large margin on both DIV2KFK and Blind144, and delivers comparable results on the DIV2KRK dataset.
It is worth noting that a simple bicubic upscaling surpasses all SotA methods in PSNR on DIV2KFK \& Blind144, indicating that these methods completely fail adapting to out-of-distribution downscaling.

Fig.~\ref{fig:sr} showcases SR results on the DIV2KFK dataset, comparing our method against SotA approaches. 
Our method demonstrates superior reconstruction quality, particularly in challenging regions such as text, structured patterns, and natural textures. Notably, in Fig.~\ref{fig:teaser}, row 1, which contains the State Bank of Mysore sign, most competing methods struggle to recover clear and legible text, often introducing excessive blurring or ghosting artifacts, whereas our method produces sharper, more readable characters. A similar trend is observed in the aerial road scene (Fig.~\ref{fig:sr}, row 4), where other methods introduce noticeable doubling artifacts (misaligned white car), which are significantly reduced in our results. Furthermore, our approach maintains a balance between sharpness and natural texture preservation in human faces and complex textures, avoiding the over-sharpening and aliasing effects present in some models (e.g., RealDAN and DIFFBIR). These results demonstrate the robustness of our approach in handling diverse realistic motion degradations present in the DIV2KFK dataset.

\begin{figure*}[t!]
    \centering
    \includegraphics[width=0.85\linewidth]{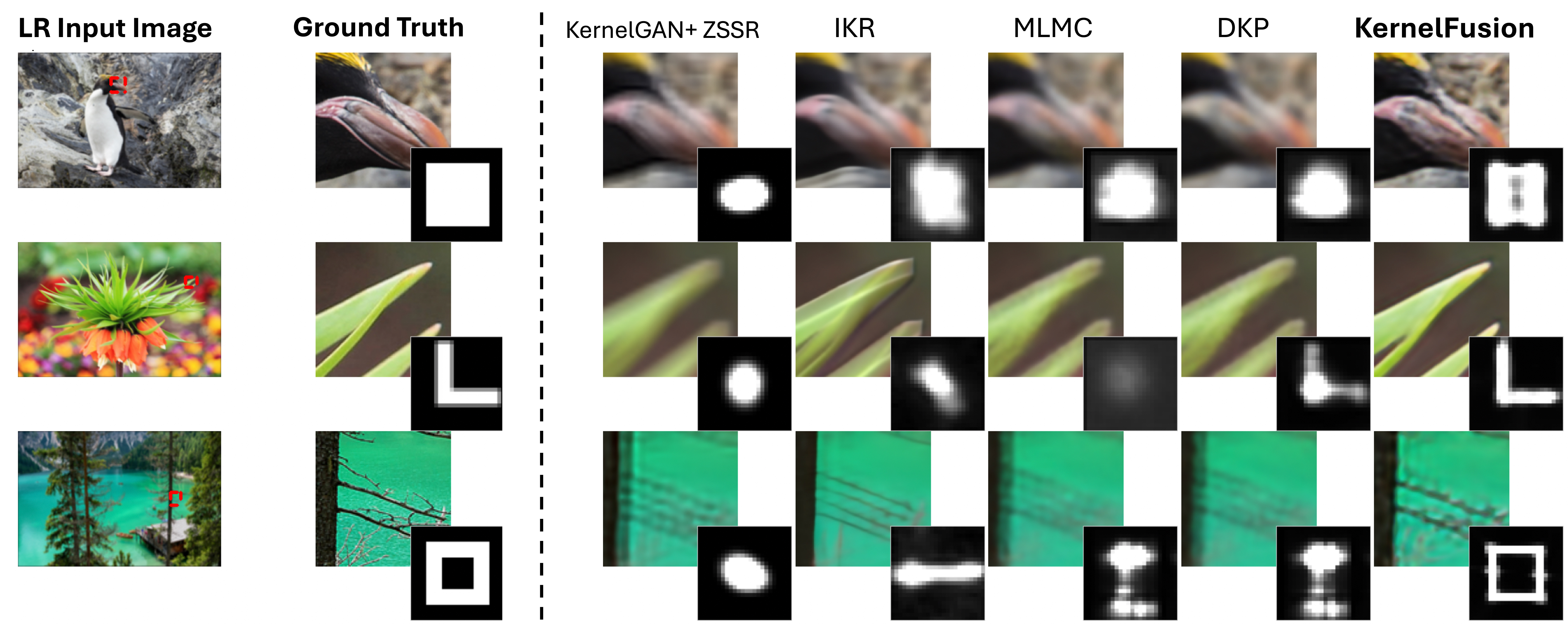}
       \caption{\small \textbf{Additional examples of explicit kernel recovery and SR under extreme downscaling kernels.}}
    \label{fig:extra}
\end{figure*}

\subsection{Kernel Evaluation}
To assess the effectiveness of different blind-SR methods in kernel estimation, we visualize the predicted  SR-kernels in Fig.~\ref{fig:est kernels vs competitors}. The top row represents the ground-truth (GT) kernels, while subsequent rows correspond to the estimated kernels from various blind-SR methods that explicitly estimate the SR-kernel (including ours).
KernelGAN demonstrates a strong bias toward Gaussian-like kernels, failing to accurately capture non-Gaussian degradations. IKR, which was trained with a mixture of random Gaussian and motion blur kernels, exhibits improved performance on motion kernels but struggles to generalize beyond its training distribution, often producing elongated blur patterns.
Both MLMC and DKP leverage meta-learning and Markov Chain Monte Carlo (MCMC) simulations to infer kernel priors without explicit assumptions. As a result, they are more flexible in capturing diverse degradation patterns. However, despite their adaptive nature, their estimated kernels still exhibit noticeable deviations from the GT, particularly in the case of extreme degradations.
Unlike prior methods, our approach does not impose any priors on the kernel shape, allowing it to accurately recover challenging degradations. 
Fig.~\ref{fig:kernel panel} presents our estimated kernels across samples from the Blind144. As observed, our approach is capable of recovering a diverse range of kernel structures, demonstrating the flexibility of our kernel estimation process. The entire collection of 144 kernels recovered by KernelFusion from the Blind144 dataset can be found in Fig.~\ref{supp:fig:blind144_kernel_panel} in the Supplementary. Additionally, Fig.~\ref{fig:extra} shows some more SR and kernel extraction comparison for extreme kernels.

\section{Limitations}
KernelFusion currently accounts for super-resolving LR images obtained under severe \emph{downscaling} degradations, which were previously considered impossible. However, while it can handle mild noise in the LR image, KernelFusion is currently not catered to handle other types of \emph{severe} degradations (like strong JPEG artifacts, or severe noise in the LR image). This is because the Patch-Diffusion competent aims to preserve the same patch statistics in the recovered HR image as in the LR input image (and would therefore include also severe noise and/or JPEG artifacts).
Extending KernelFusion to handle other types of degradations \emph{in a blind manner} is part of our future work.

Additionally, our method relies solely on the internal statistics of a single image (the LR image), without leveraging any external priors or pre-trained large-scale diffusion models. Although this assumption-free design is central to our ability to tackle previously intractable downscaling degradations, it 
does not exploit rich external information which could enhance reconstruction quality. 
Moreover, patch diffusion has to be trained from scratch on each new LR image, taking $\sim20$ minutes per image. The following upscaling process then depends on input image size and scale factor applied.
Exploring approaches to integrate \emph{external} learned priors with our \emph{internal} Patch-Diffusion model may significantly speed up KernelFusion, as well as potentially improve its output quality. This represents an exciting direction for future research.

\section{Conclusion}
``KernelFusion'' recovers the unique image-specific SR-kernel directly from the LR input image, while simultaneously recovering its corresponding HR image in a consistent manner.
It can handle complex downscaling degradations, where existing SotA Blind-SR methods fail miserably. Being a zero-shot estimation method which trains \emph{internally} on the LR input image only, KernelFusion is not bound by any external training  distribution, hence can handle any type of downscaling kernel. It has no notion of ``out-of-distribution'' kernels, which \emph{externally-trained} Blind-SR methods suffer from.
By breaking free from predefined kernel assumptions, KernelFusion pushes Blind-SR into a new, assumption-free paradigm — handling downscaling degradations previously thought impossible.
\clearpage 

{
    \small
    \bibliographystyle{ieeenat_fullname}
    \bibliography{main}
}

\clearpage  % Ensures the supplementary material starts on a new page
\appendix   % Makes section numbers appear as A1, A2, etc.

\setcounter{section}{0}  % Reset section counter
\renewcommand{\thesection}{A\arabic{section}}  % Sections as A1, A2, ...

\setcounter{figure}{0}  % Reset figure counter
\renewcommand{\thefigure}{S\arabic{figure}}  % Figures as S1, S2, ...

\setcounter{table}{0}  % Reset table counter (optional)
\renewcommand{\thetable}{ST\arabic{table}}  % Tables as ST1, ST2, ...

\twocolumn[{%
    \centering
    \LARGE \textbf{Supplementary Material} \par  % Large, bold, centered 
    \vspace{1.em}  % Space below title
  \hspace*{-1cm}  \includegraphics[width=1.1\textwidth]{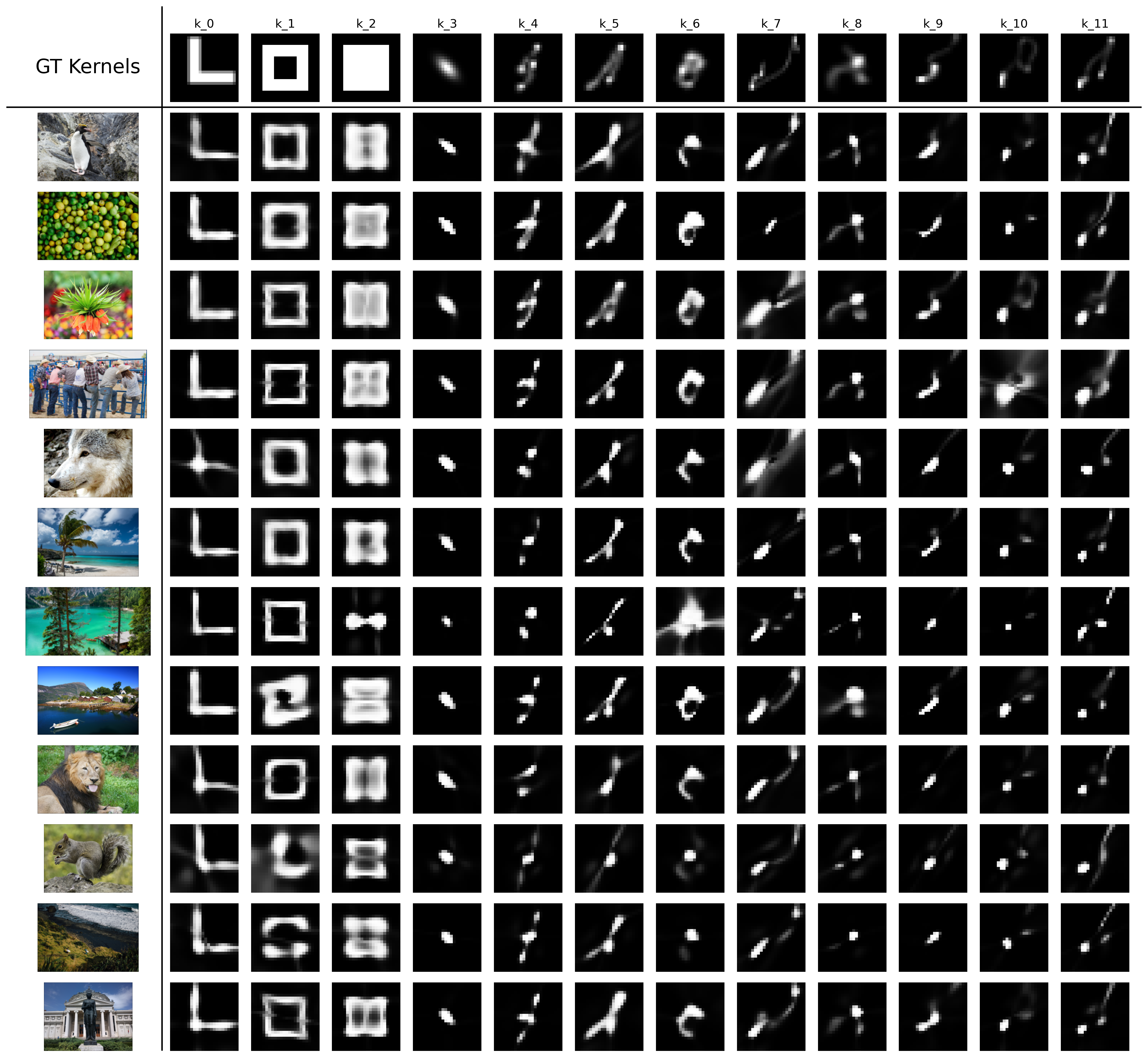}
    \captionof{figure}{\textbf{Estimated kernels of KernelFusion on Blind144:} Complete overview of all 144 estimated kernels.}
    \label{supp:fig:blind144_kernel_panel}
}]
\clearpage 

\section{Further Technical Details}
\label{supp:sec:technical details}
\subsection{Patch-Diffusion}
\paragraph{Architecture}
The backbone model is a convolutional network, that inputs an image tensor $x$ and a timestep $t$. We have a total of 6 blocks, each block conditioned on $t$. We use one block of two 3 $\times$ 3 filters, followed by six blocks of 3 $\times$ 3 + 1 $\times$ 1 filters. We use 128 filters for the hidden layers.

\paragraph{Training Details}
The patch diffusion model is trained using random crops of size 64 pixels. The model is trained for $600'000$ steps, using Adam as optimizer, with a learning rate of $lr=1\times10^{-4}$ and cosine annealing.

\subsection{U-Net}
\label{supp:subsec:unet}
\paragraph{Architecture}
The refinement UNET consists of 5 blocks, with 32 filters on the input level and 512 filters on the bottom level.
Each block consists of 2 convolutional layers with a 3$\times$3 kernel, ReLU activations and batch norm. The final layer uses a tanh activation function to ensure that the predicted $x_0$ output is in the expected -1 to 1 range.
The levels are down respectively upsampled by a scale factor of 2.

\paragraph{Training Details}
\label{supp:par:training details unet}
The UNET is trained at each time step $t$ during the reverse diffusion process. We use Adam optimizer, a learning rate of $lr=1\times10^{-4}$. We apply cosine annealing, reducing the learning rate at each $t$ to $lr=5\times10^{-5}$. The UNET is initialized at the first $T_{start}$ of the reverse diffusion process and then finetuned along the different timesteps $t$. With the exception of the initial $T_{start}$ where we apply 100 iterations, the model is then finetune for 20 iterations during each $t$.

\subsection{Implicit Neural Representation for Kernel Estimation}
\paragraph{Architecture}
As described in Sec.~\ref{subsec:kernel estimation}, we took inspiration from SIREN \cite{sitzmann2020implicit} for our implicit neural representation. The network consists of 5 fully connected layers, with 256 nodes each. In contrast to the original paper, we reduced $\omega$ from 30 to 5 and apply it across all layers. Our last layer has an activation function which we call \emph{leaky sigmoid}, a sigmoid function also allowing for slight negative values: $\sigma_{leaky}(x)=(1+10^{-4})\cdot\sigma(x)-10^{-4}$.

\paragraph{Training Details}
As we train our INR along with the UNET, we use the same training setup as described in Sec.~\ref{supp:subsec:unet}.

\end{document}